\documentclass[letterpaper,10pt,conference]{ieeeconf}
\usepackage{amsmath,amssymb,amsfonts}
\usepackage{booktabs}
\usepackage{multirow}
\usepackage{cite}
\usepackage{graphicx}
\usepackage{microtype}
\usepackage{tikz}
\usepackage{url}
\usetikzlibrary{arrows.meta,backgrounds,calc,positioning}
\newcommand{\method}{FlashNeRD}
\title{\LARGE \bf FlashNeRD: Performance-First Contact-Rich Neural Robot Dynamics}
\author{
  Mohammadmehdi Ataei, Mohammad Amin Nabian, Jie Xu, Miles Macklin,
  Yashraj Narang, \\
  Eric Heiden, Octi Zhang, Yunrong Guo, Ruben D'Sa, Oliver Hennigh, Gavriel State, and Sanjay Choudhry \\[6pt]
  {\itshape NVIDIA}
}
\begin{document}
\maketitle
\thispagestyle{empty}
\pagestyle{empty}

\begin{abstract}
    Compared with analytical physics, learned dynamics models promise robot simulation that is faster, inherently differentiable, and easily adaptable to real data. Neural Robot Dynamics (NeRD) pursues this by keeping collision detection analytical and replacing a simulator's numerical dynamics for the robot with a learned model. Three limitations remain. NeRD offers little speedup over the simulator it learned from, accepts contact only at predefined points, and has no two-way coupling with objects it manipulates. FlashNeRD removes all three with a parallel streaming architecture that makes each prediction faster and more accurate, an encoder that accepts contacts wherever they occur, and two-way coupling with objects simulated by analytical solvers. Experiments across five robots show faster and more accurate dynamics, faster policy learning, and faster inference-time planning. Across three robots, FlashNeRD's dynamics model is up to $55\times$ faster than an optimized NeRD and more accurate over long rollouts. This speedup extends to policy learning, where PPO trains an ANYmal locomotion policy in 34 seconds, $3.8\times$ faster than the analytical simulator and $2.7\times$ faster than an optimized NeRD. With DIAL-MPC, a sampling-based MPC method, the robot climbs all six test platforms where fixed-contact NeRD manages one, at approximately half the analytical simulator's planning time. Cube-reorientation policies trained with FlashNeRD complete within 2\% of the simulator-trained policy's target count.
\end{abstract}

\section{Introduction}
Robot learning and planning require many dynamics predictions, making simulation cost a practical bottleneck. Analytical simulators such as MuJoCo \cite{todorov2012mujoco} and GPU-parallel platforms such as Isaac Lab \cite{mittal2025isaaclab} and Newton \cite{mujocowarp}, whose MuJoCo Warp (MJWarp) backend we use unless otherwise stated, meet this demand by stepping thousands of environments. Three limits persist. Calls to an iterative contact solver bound throughput per GPU, many simulators are not differentiable, and fitting a simulator to real-robot data means tuning a few physical parameters that cannot express unmodeled effects such as actuator dynamics.

Neural robot dynamics learns the robot's transition function from data, addressing all three limits in principle. Batched network evaluation has fixed cost and can outpace iterative contact solves. The model is differentiable by construction. The model can also be fine-tuned on real trajectories without selecting physical parameters to adjust. Neural Robot Dynamics (NeRD) \cite{pmlr-v305-xu25g} is the state of the art, using one transformer per robot, trained on randomized states and actuator commands such as position targets or motor torques, with collision detection left analytical. Because the data are task- and controller-independent, the model learns generic robot dynamics rather than task-specific dynamics and serves any controller on the robot. NeRD also showed zero-shot sim-to-real transfer on reach tasks.

Three limitations remain. (i) NeRD re-encodes the full history window at every step, so inference cost is similar to the analytical simulator's and the throughput advantage goes unrealized. (ii) Contact enters through predefined probe points, which are insufficient for more complex tasks where contacts happen at edges or surfaces, such as climbing. (iii) NeRD assumes the environment surrounding the robot is fully static, ruling out manipulation. \method{} removes each limitation, and each fix unlocks a capability.
\begin{itemize}
\item \textbf{A parallel architecture with streaming inference} that processes only the newest frame per step. This gives up to $55\times$ the model-only throughput of an inference-optimized NeRD baseline (\emph{Optimized NeRD}) with lower rollout error, and is fast enough that end-to-end policy training also runs faster than in the analytical simulator.

\item \textbf{A set-based contact encoder} that accepts contacts wherever collisions are detected, so contact-rich tasks such as platform climbing need no hand-placed probes.
\item \textbf{Two-way coupling via hybrid simulation}, in which \method{} predicts the robot's free motion while MJWarp analytically simulates the objects and resolves their contacts with the robot, so forces flow in both directions and learned dynamics extend to manipulation.
\end{itemize}
We validate each capability across five robots, spanning policy learning and inference-time planning with the learned model. \method{} trains a locomotion policy $3.2\times$ faster than the analytical simulator. With sampling-based MPC, the robot climbs all six test platforms where NeRD manages one, and a cube-reorientation policy trained with \method{} completes within $2\%$ of the simulator-trained policy's target count.

\section{Related Work}
Neural simulation has been explored for fluids~\cite{sanchezgonzalez2020gns}, deformable bodies~\cite{pfaff2021learning, bertiche2022cloth}, and rigid bodies~\cite{sanchez2018graph, li2025rwm, zhang2026nrd, dou2026rigidformerlearningrigiddynamics, andriluka2024neural, heiden2021neuralsim}. Among neural simulators for articulated rigid bodies, NeRD~\cite{pmlr-v305-xu25g} learns robot-specific physics from robot-centric states, controls, and contact features independent of downstream tasks. FlashNeRD builds on it with higher inference speed, variable contacts, and two-way coupling. Three bodies of work inform the design. Efficient transformer inference, including parallel attention and feedforward branches~\cite{chowdhery2023palm}, multi-query attention~\cite{shazeer2019fast}, sliding-window attention~\cite{jiang2023mistral}, IO-aware kernels~\cite{dao2022flashattention}, and object-level attention in RigidFormer~\cite{dou2026rigidformerlearningrigiddynamics}, motivates the architecture of Secs.~\ref{sec-arch} and \ref{sec-cache}. In place of contact representations such as implicit signed distance functions~\cite{pmlr-v155-pfrommer21a}, mesh-face graphs~\cite{allen2023learning}, and predicted binary contacts~\cite{jing2026contactawareneuraldynamics}, FlashNeRD pools the analytical detector's unordered contact set. Coupling schemes between solvers~\cite{shinar2008twoway, chen2026divide} or learned models~\cite{dou2025cfc} motivate the shared contact solve of Sec.~\ref{sec-coupling-method}.

\section{Method}
\label{sec-method}

FlashNeRD has three components, one per limitation above, all built on the shared transition-learning problem of Sec.~\ref{sec-transition}. The first is a transformer that evaluates attention and feedforward branches in parallel within each block and caches past keys and values, so each rollout step encodes only the newest frame (Secs.~\ref{sec-arch} and \ref{sec-cache}). The second is a contact encoder that accepts contacts wherever the collision detector finds them (Sec.~\ref{sec-variable-method}). The third is two-way coupling via hybrid simulation, in which analytically simulated objects push back on the learned robot (Sec.~\ref{sec-coupling-method}).

\subsection{Learning a Robot Transition}
\label{sec-transition}

A robot state $s_t$ contains joint positions and velocities plus, for floating-base robots, base orientation and linear and angular velocity. Each input frame $x_t=(s_t,u_t,\vec g_t)$ adds control $u_t$ and gravity direction $\vec g_t$. Contact input $\mathcal C_t$ is either an ordered, fixed-length vector from predefined link probes queried against collision geometry each step or a variable set of detected contacts. Fixed contact suits known locations such as feet on flat ground, while variable contact handles changing locations such as edges. The model predicts a state increment from $R$ frames, including the current frame,
\begin{equation}
 \widehat{\Delta s}_{t+1}
 = F_\theta(x_{t-R+1},\ldots,x_t,\mathcal C_{t-R+1},\ldots,\mathcal C_t),
 \label{eq-transition}
\end{equation}
where hats denote predictions. We add the increment to $s_t$ and compose orientation increments as rotations. Expressing $s_t$, $\vec g_t$, $\mathcal C_t$, and $\widehat{\Delta s}_{t+1}$ in the robot's base frame removes dependence on world position and heading.

\begin{figure*}[t]
 \centering
 \resizebox{0.9\textwidth}{!}{\begingroup
\definecolor{nvGreen}{HTML}{76B900}
\definecolor{nvGreenDark}{HTML}{4C8500}
\definecolor{nvInk}{HTML}{1A1D1A}
\definecolor{nvMuted}{HTML}{626862}
\definecolor{nvPanelEdge}{HTML}{C8CEC4}
\definecolor{nvGreenFill}{HTML}{EDF6E2}
\begin{tikzpicture}[
 x=1cm,y=1cm,font=\fontfamily{cmss}\fontsize{7}{8.4}\selectfont,line cap=round,line join=round,
 flow/.style={-{Stealth[length=1.35mm,width=1.05mm]},draw=nvInk!76,line width=0.64pt,shorten <=0.5pt,shorten >=0.8pt,rounded corners=0.8mm},
 contact flow/.style={flow,draw=nvGreenDark,line width=0.74pt},
 wire/.style={draw=nvInk!76,line width=0.64pt,rounded corners=0.8mm},
 contact wire/.style={wire,draw=nvGreenDark,line width=0.74pt},
 box/.style={draw=nvInk!52,fill=white,rounded corners=1.05mm,line width=0.6pt,align=center,inner xsep=1.1mm,inner ysep=0.95mm,minimum height=5.6mm},
 learned/.style={box,draw=nvGreenDark,line width=0.7pt},
 sum/.style={circle,draw=nvGreenDark,fill=white,line width=0.7pt,minimum size=3.8mm,inner sep=0pt,font=\fontfamily{cmss}\fontsize{8}{9.6}\selectfont\bfseries},
 title/.style={anchor=west,font=\fontfamily{cmss}\fontsize{8}{9.6}\selectfont\bfseries,text=nvInk},
 note/.style={fill=white,inner sep=1pt,text=nvMuted},
 token/.style={draw=nvGreenDark,fill=nvGreenFill,rounded corners=0.35mm,line width=0.5pt,minimum width=5.8mm,minimum height=2.6mm,inner sep=0pt},
 padding/.style={token,draw=nvPanelEdge,fill=white,densely dashed},
 feature/.style={circle,draw=nvGreenDark,fill=white,line width=0.55pt,minimum size=3.7mm,inner sep=0pt},
 slot/.style={draw=nvGreenDark,fill=nvGreenFill,rounded corners=0.4mm,line width=0.6pt,minimum width=6.8mm,minimum height=3.9mm,inner sep=0pt},
 pool edge/.style={draw=nvGreenDark!45,line width=0.45pt}]
\path[use as bounding box] (0,0) rectangle (18.1,6.0);
\path[draw=nvPanelEdge,fill=white,line width=0.5pt,rounded corners=1.6mm]
 (0.05,0.05) rectangle (18.05,5.95);
\draw[draw=nvPanelEdge,line width=0.5pt] (6.13,0.55)--(6.13,5.45);
\draw[draw=nvPanelEdge,line width=0.5pt] (13.65,0.55)--(13.65,5.45);
\fill[nvGreen] (0.26,5.48) rectangle (0.34,5.76);
\node[title] at (0.44,5.62) {(a) Variable-contact encoding};
\fill[nvGreen] (6.29,5.48) rectangle (6.37,5.76);
\node[title] at (6.47,5.62) {(b) Parallel temporal block ($\times L$)};
\fill[nvGreen] (13.81,5.48) rectangle (13.89,5.76);
\node[title] at (13.99,5.62) {(c) Increment head};

\node[note,text=nvInk] at (1.8,5.13) {Simulator-generated contacts};
\path[draw=nvGreenDark!55,fill=nvGreen!7,line width=0.5pt,rounded corners=0.9mm]
 (1.34,3.51) rectangle (2.26,4.91);
\foreach \x/\time in {0.75/{t-1},1.8/{t},2.85/{t+1}}
 {\node[note] at (\x,4.7) {$\time$};}
\node[token] at (0.75,4.36) {};
\node[token] at (0.75,4.07) {};
\node[padding] at (0.75,3.78) {};
\node[token] at (1.8,4.36) {};
\node[token] at (1.8,4.07) {};
\node[token] at (1.8,3.78) {};
\node[token] at (2.85,4.36) {};
\node[padding] at (2.85,4.07) {};
\node[padding] at (2.85,3.78) {};
\node[note,font=\fontfamily{cmss}\fontsize{5}{6}\selectfont] at (0.75,3.47) {Masked};
\draw[draw=nvPanelEdge,line width=0.5pt] (0.75,3.56)--(0.75,3.64);
\node[learned,minimum width=2.92cm] (encoder) at (1.8,2.84) {Shared row encoder $\phi(c_{t,j})$};
\draw[contact flow] (1.8,3.48)--node[note,right=1mm,text=nvGreenDark] {$\mathcal C_t$} (encoder.north);
\node[feature] (e1) at (0.8,2.18) {$e_1$};
\node[feature] (e2) at (1.8,2.18) {$e_2$};
\node[feature] (en) at (2.8,2.18) {$e_{n_t}$};
\node[note,font=\fontfamily{cmss}\fontsize{5}{6}\selectfont,inner sep=0pt] at (2.28,2.18) {$\cdots$};
\draw[contact flow] ([xshift=-1cm]encoder.south)--(e1.north);
\draw[contact flow] (encoder.south)--(e2.north);
\draw[contact flow] ([xshift=1cm]encoder.south)--(en.north);

\path[draw=nvGreenDark!55,fill=white,line width=0.55pt,rounded corners=0.8mm]
 (0.38,0.58) rectangle (4.96,1.56);
\foreach \from in {e1,e2,en}{
 \draw[pool edge] (\from.south)--(0.95,1.37);
 \draw[pool edge] (\from.south)--(1.85,1.37);
 \draw[pool edge] (\from.south)--(3.1,1.37);
}
\node[note,text=nvGreenDark,anchor=west,align=left] at (3.25,1.94) {Learned set\\pooling};
\node[slot] at (0.95,1.16) {$b_1$};
\node[slot] at (1.85,1.16) {$b_2$};
\node[slot] at (3.1,1.16) {$b_M$};
\node[note,font=\fontfamily{cmss}\fontsize{5}{6}\selectfont,inner sep=0pt] at (2.48,1.16) {$\cdots$};
\draw[draw=nvPanelEdge,line width=0.5pt] (3.61,0.76)--(3.61,1.38);
\node[slot,fill=white] at (4.25,1.16) {$\rho_t$};
\node[note] at (2.0,0.76) {$M$ slots};
\node[note] at (4.25,0.76) {Count};
\coordinate (bankEast) at (4.96,1.16);
\coordinate (bankSouth) at (2.67,0.58);

\node[box,minimum width=1.95cm,minimum height=7.4mm] (frame) at (4.5,4.94)
 {Current frame\\$x_t$};
\node[learned,minimum width=1.48cm] (projection) at (4.5,3.9) {Input\\projection};
\node[sum] (inputsum) at (5.81,3.9) {$+$};
\draw[flow] (frame.south)--(projection.north);
\draw[flow] (projection.east)--(inputsum.west);
\node[learned,fill=nvGreenFill,minimum width=5mm,minimum height=3.8mm,inner xsep=0.5mm,inner ysep=0.3mm] (contactin) at (5.81,2.42) {$G_{\mathrm{in}}$};
\draw[contact wire] (bankEast)--(5.81,1.16)--(contactin.south);
\draw[contact flow] (contactin.north)--(inputsum.south);

\coordinate (fork) at (6.22,3.9);
\node[box,minimum width=0.96cm,inner xsep=0.5mm] (rms) at (7.1,3.9) {RMSNorm};
\node[learned,minimum width=1.2cm,minimum height=12mm,inner xsep=0.5mm,right=2.5mm of rms] (packed)
 {Packed\\projection\\$W_{\mathrm{in}}$};
\coordinate (split) at (9.3,3.9);
\node[learned,minimum width=1.7cm,minimum height=7.4mm,inner xsep=0.6mm] (attention) at (10.5,4.37)
 {MQA + \\Relative bias};
\node[box,minimum width=1.7cm,minimum height=7.4mm,inner xsep=0.6mm] (ffn) at (10.5,3.43)
 {Feedforward\\SwiGLU};
\node[learned,minimum width=2.6cm,minimum height=4mm,inner ysep=0.5mm] (cache) at (10.5,5.2)
 {Bounded KV cache};
\node[learned,minimum width=1.15cm,minimum height=15mm,inner xsep=0.5mm] (joined) at (12.25,3.9)
 {Concat.\\+ project\\$W_{\mathrm{out}}$};
\node[sum] (blocksum) at (13.4,3.9) {$+$};
\draw[wire] (inputsum.east)--(fork);
\fill[nvInk!76] (fork) circle[radius=0.62pt];
\draw[flow] (fork)--(rms.west);
\draw[flow] (rms.east)--(packed.west);
\draw[wire] (packed.east)--(split);
\draw[wire] (9.3,4.37)--(9.3,3.43);
\fill[nvInk!76] (split) circle[radius=0.62pt];
\draw[flow] (9.3,4.37)--(attention.west);
\draw[flow] (9.3,3.43)--(ffn.west);
\draw[flow] (attention.east)--(attention.east -| joined.west);
\draw[flow] (ffn.east)--(ffn.east -| joined.west);
\draw[flow] (cache.south)--(attention.north);
\draw[flow] (joined.east)--(blocksum.west);
\draw[flow] (fork)--(6.22,2.55)--(13.4,2.55)--(blocksum.south);
\node[note] at (9.3,2.55) {Residual $h_t^{\ell-1}$};

\node[box,minimum width=1.7cm,minimum height=7.2mm] (normhead) at (15.1,4.87)
 {RMSNorm\\$h_t^L$};
\node[box,minimum width=0.7cm,minimum height=5.6mm] (current) at (17.12,4.87) {$x_t$};
\node[learned,minimum width=1.9cm,minimum height=7.2mm] (prehead) at (15.1,3.79)
 {Concat. + affine\\$W_h(\cdot)+b_h$};
\node[sum] (readsum) at (15.1,2.97) {$+$};
\node[learned,minimum width=1.8cm,minimum height=7.2mm] (head) at (15.1,2.23)
 {ReLU + affine\\$W_o(\cdot)+b_o$};
\node[learned,minimum width=1.4cm,minimum height=7.2mm] (linear) at (17.12,2.23)
 {Affine path\\$W_p x_t+b_p$};
\node[sum] (predsum) at (15.1,1.48) {$+$};
\node[learned,minimum width=2.14cm,minimum height=8mm] (output) at (15.1,0.59)
 {Predicted increment\\[0.55ex]$\widehat{\Delta s}_{t+1}$};
\draw[flow] (blocksum.east)--(13.92,3.9)--(13.92,4.87)--(normhead.west);
\draw[flow] (normhead.south)--(prehead.north);
\draw[flow] (current.west)--(16.57,4.87)--(16.57,3.79)--(prehead.east);
\draw[flow] (prehead.south)--(readsum.north);
\draw[flow] (readsum.south)--(head.north);
\draw[flow] (head.south)--(predsum.north);
\draw[flow] (predsum.south)--(output.north);
\draw[flow] (current.south)--(linear.north);
\draw[flow] (linear.south)--(17.12,1.48)--(predsum.east);
\node[learned,fill=nvGreenFill,minimum width=6mm,minimum height=3.8mm,inner xsep=0.5mm,inner ysep=0.3mm] (contactread) at (14.22,2.97) {$G_{\mathrm r}$};
\node[learned,fill=nvGreenFill,minimum width=6mm,minimum height=3.8mm,inner xsep=0.5mm,inner ysep=0.3mm] (contactpred) at (14.22,1.48) {$G_{\mathrm p}$};
\draw[contact wire] (bankSouth)--(2.67,0.3)--(13.75,0.3)--(13.75,2.97)--(contactread.west);
\node[note,text=nvGreenDark] at (9.1,0.3) {Shared contact summary $(B_t,\rho_t)$};
\fill[nvGreenDark] (13.75,1.48) circle[radius=0.62pt];
\draw[contact wire] (13.75,1.48)--(contactpred.west);
\draw[contact flow] (contactread.east)--(readsum.west);
\draw[contact flow] (contactpred.east)--(predsum.west);
\end{tikzpicture}
\endgroup}
\caption{FlashNeRD overview (variable-contact case). (a) A shared row encoder and learned pooling compress detected contacts into $M$ fixed-size summary slots plus a count, accepting contacts anywhere on the robot. (b) One of $L$ blocks. Attention reads the current frame and a short cache of past keys and values, while a gated feedforward branch transforms the current frame. Both run in parallel from one packed projection, processing only the newest frame per step. (c) Readout. Temporal features, the current frame, and a direct affine path predict the state increment, retaining current-state information. Green outlines mark learned modules, and green arrows carry contact information.}
\label{fig-architecture}
\end{figure*}
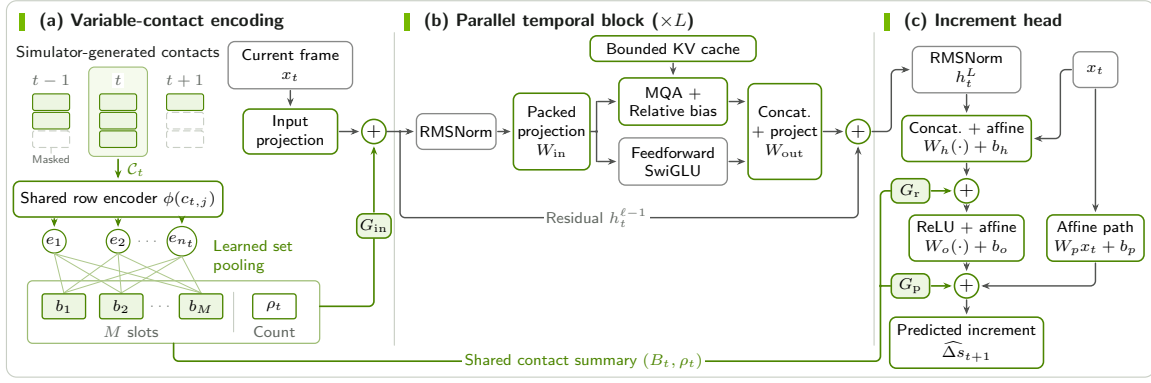

Following NeRD \cite{pmlr-v305-xu25g}, we train \method{} on randomized physical states and actuator commands independently of any controller, retaining analytical collision detection. Each robot's model serves RL and planning with the same actuator model and timestep, provided training covers the relevant states and contacts. Figure~\ref{fig-architecture} gives an overview.

\subsection{An Architecture for Fast and Accurate Rollouts}
\label{sec-arch}

Two considerations shape the design. First, autoregressive rollout time grows approximately with horizon times per-step latency. Kernel overhead and memory traffic can substantially affect this latency, especially at small batch sizes, motivating reductions in sequential operations and data movement alongside arithmetic cost. Second, a short history provides temporal context that can help the model distinguish consistent motion patterns from noise or small errors in individual frames. Combining information across recent frames can therefore make predictions less sensitive to fluctuations in any single input state. The design retains this temporal context while minimizing sequential work per step. Section~\ref{sec-cache} then removes NeRD's repeated encoding of the full history window.

Each of $L$ blocks has two branches that read the same normalized input and run in parallel, following PaLM \cite{chowdhery2023palm} rather than the serial attention-then-feedforward layout of a standard transformer \cite{vaswani2017attention}. Attention reads the current and $S_\ell-1$ preceding frames and captures history-dependent effects, while the feedforward branch transforms the current hidden state. The embedding is $h_t^0=W_e x_t+b_e+G_{\mathrm{in}}(\mathcal C_t)$, where $G_{\mathrm{in}}$ is a linear map of the contact probe vector for fixed-contact models and a gated map of the pooled contact bank (Sec.~\ref{sec-variable-method}) for variable-contact models. Block $\ell$ computes
\begin{equation}
\begin{aligned}
z_t^\ell
    &= \operatorname{RMSNorm}(h_t^{\ell-1}),\\
[q_t^\ell,k_t^\ell,v_t^\ell,g_t^\ell,r_t^\ell]
    &= W_{\mathrm{in}}^\ell z_t^\ell,\\
a_t^\ell
    &= \underbrace{
       \operatorname{MQA}_{S_\ell}
       (q_t^\ell,k_{\leq t}^\ell,v_{\leq t}^\ell)
       }_{\text{\scriptsize temporal-attention branch}},\\
f_t^\ell
    &= \underbrace{
       \operatorname{SiLU}(g_t^\ell)\odot r_t^\ell
       }_{\text{\scriptsize feedforward branch}},\\
h_t^\ell
    &= h_t^{\ell-1}
       + W_{\mathrm{out}}^\ell[a_t^\ell,f_t^\ell],
\end{aligned}
\label{eq-block}
\end{equation}
where square brackets denote concatenation, RMSNorm \cite{zhang2019rmsnorm} normalizes the input, and the feedforward branch is a SwiGLU gate \cite{shazeer2020glu}. Multi-query attention (MQA) \cite{shazeer2019fast} uses one key/value head for all query heads, so the cache read per step stays small. Both branches are produced by one packed input projection and merged by one output projection, halving the sequential launches per block relative to the serial layout.

Temporal order is encoded by a learned relative positional bias. A learned scalar indexed by lag $t-j$ is added to each query-key score before the softmax, with lag zero the current frame. Because the bias depends only on lag, it is unaffected by where a frame sits in the circular cache of Sec.~\ref{sec-cache}.

Attention summarizes history, but the increment depends most on the current configuration, velocity, and control. To retain this information, the readout combines a nonlinear head with a direct affine path from the current frame,
\begin{equation}
\begin{aligned}
 p_t &= W_h[\operatorname{RMSNorm}(h_t^L),x_t]+b_h
       +G_{\mathrm r}(\mathcal C_t),\\
 \widehat{\Delta s}_{t+1}
     &= W_o\operatorname{ReLU}(p_t)+b_o\\
     &\quad+W_p x_t+b_p+G_{\mathrm p}(\mathcal C_t),
\end{aligned}
\label{eq-readout}
\end{equation}
where $G_{\mathrm r}$ and $G_{\mathrm p}$ inject contact information into the hidden readout and the increment (Sec.~\ref{sec-variable-method}) and are omitted for fixed-contact models. The affine path captures approximately linear local dynamics, and the nonlinear head adds state-dependent corrections. The direct terms are initialized to zero so training starts from the nonlinear predictor alone.

Inputs, outputs, and contact rows are standardized with NeRD's normalization (omitted from the equations), and the model is trained by mean squared error on standardized one-step increments with invalid transitions and padding masked. Beyond reducing sequential work, the parallel form decouples temporal-context modeling from current-feature transformation: attention captures history-dependent effects, while the feedforward branch processes current features without first passing through the same block's attention. This decoupling, together with the direct current-frame readout, has been shown in our controlled experiments to improve prediction accuracy and reduce recursive rollout drift.

\subsection{Streaming Dynamics with Bounded KV Caching}
\label{sec-cache}

A full-window transformer re-embeds and re-projects all $R$ frames at every step even though $R-1$ of them were processed at the previous step. In FlashNeRD only the newest frame is projected, and the keys and values of earlier frames are reused from a cache. The reuse is deliberately bounded. No block attends beyond its span in training, and lag biases exist only within the window, so an unbounded cache would renormalize attention over untrained key/value pairs and lags, and predictions would no longer be those of the trained model.

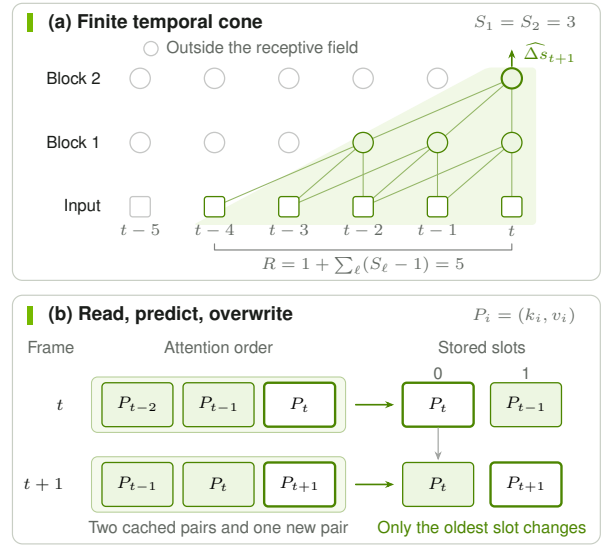
\begin{figure}[t]
 \centering
 \resizebox{\columnwidth}{!}{\begingroup
\definecolor{nvGreen}{HTML}{76B900}
\definecolor{nvGreenDark}{HTML}{4C8500}
\definecolor{nvInk}{HTML}{1A1D1A}
\definecolor{nvMuted}{HTML}{626862}
\definecolor{nvPanelEdge}{HTML}{C8CEC4}
\definecolor{nvGreenFill}{HTML}{EDF6E2}
\begin{tikzpicture}[
 x=1cm,y=1cm,font=\sffamily\scriptsize,line cap=round,line join=round,
 panel/.style={draw=nvPanelEdge,fill=white,line width=0.5pt,rounded corners=1.5mm},
 title/.style={anchor=west,font=\sffamily\footnotesize\bfseries,text=nvInk},
 label/.style={text=nvMuted,inner sep=1pt},
 row label/.style={anchor=east,align=right,text=nvInk},
 frame/.style={rectangle,draw=nvGreenDark,fill=white,rounded corners=0.4mm,line width=0.55pt,minimum size=3mm,inner sep=0pt},
 feature/.style={circle,draw=nvGreenDark,fill=nvGreenFill,line width=0.55pt,minimum size=3mm,inner sep=0pt},
 outside/.style={draw=nvInk!25,fill=white},
 cone edge/.style={draw=nvGreenDark!60,line width=0.45pt},
 flow/.style={-{Stealth[length=1.35mm,width=1.05mm]},draw=nvGreenDark,line width=0.7pt,shorten >=0.6pt},
 write/.style={flow,densely dashed,rounded corners=0.9mm},
 cell/.style={draw=nvGreenDark,fill=nvGreenFill,rounded corners=0.6mm,line width=0.55pt,minimum width=12mm,minimum height=6.5mm,inner sep=0pt}]
\path[use as bounding box] (0,0) rectangle (8.8,8.1);
\path[panel] (0.05,3.95) rectangle (8.75,8.05);
\path[panel] (0.05,0.05) rectangle (8.75,3.73);
\fill[nvGreen] (0.27,7.62) rectangle (0.35,7.9);
\node[title] at (0.46,7.76) {(a) Finite temporal cone};
\node[label,anchor=east] at (8.43,7.76) {$S_1=S_2=3$};
\fill[nvGreen!10,rounded corners=0.6mm]
 (7.08,7.1)--(7.8,7.1)--(7.8,4.77)--(2.72,4.77)--cycle;
\foreach \a in {5.24,6.34,7.44}{\draw[cone edge] (7.44,6.94)--(\a,5.99);}
\foreach \a in {3.04,4.14,5.24}{\draw[cone edge] (5.24,5.99)--(\a,5.04);}
\foreach \a in {4.14,5.24,6.34}{\draw[cone edge] (6.34,5.99)--(\a,5.04);}
\foreach \a in {5.24,6.34,7.44}{\draw[cone edge] (7.44,5.99)--(\a,5.04);}
\node[frame,outside] at (1.94,5.04) {};
\foreach \a in {3.04,4.14,5.24,6.34,7.44}{\node[frame] at (\a,5.04) {};}
\foreach \a in {1.94,3.04,4.14}{\node[feature,outside] at (\a,5.99) {};}
\foreach \a in {5.24,6.34,7.44}{\node[feature] at (\a,5.99) {};}
\foreach \a in {1.94,3.04,4.14,5.24,6.34}{\node[feature,outside] at (\a,6.94) {};}
\node[feature,line width=0.9pt] at (7.44,6.94) {};
\node[row label] at (1.48,6.94) {Block 2};
\node[row label] at (1.48,5.99) {Block 1};
\node[row label] at (1.48,5.04) {Input};
\foreach \a/\age in {1.94/5,3.04/4,4.14/3,5.24/2,6.34/1}
 {\node[label] at (\a,4.68) {$t-\age$};}
\node[label] at (7.44,4.68) {$t$};
\draw[draw=nvInk!55,line width=0.5pt] (3.04,4.49)--(3.04,4.4)--(7.44,4.4)--(7.44,4.49);
\node[label] at (5.24,4.17) {$R=1+\sum_\ell(S_\ell-1)=5$};
\node[feature,outside,minimum size=2.1mm] at (2.1,7.38) {};
\node[label,anchor=west] at (2.29,7.38) {Outside the receptive field};
\draw[flow] (7.44,7.12)--(7.44,7.39);
\node[label,anchor=west,text=nvGreenDark] at (7.57,7.31) {$\widehat{\Delta s}_{t+1}$};

\fill[nvGreen] (0.27,3.29) rectangle (0.35,3.57);
\node[title] at (0.46,3.43) {(b) Read, predict, overwrite};
\node[label,anchor=east] at (8.43,3.43) {$P_i=(k_i,v_i)$};
\node[label] at (0.62,2.96) {Frame};
\node[label] at (3.1,2.96) {Attention order};
\node[label] at (7.0,2.96) {Stored slots};
\node[label] at (6.35,2.57) {0};
\node[label] at (7.65,2.57) {1};
\node[row label] at (0.93,2.11) {$t$};
\node[row label] at (0.93,0.93) {$t+1$};
\path[draw=nvGreenDark!40,fill=nvGreen!7,rounded corners=0.9mm,line width=0.45pt]
 (1.22,1.7) rectangle (4.98,2.52);
\path[draw=nvGreenDark!40,fill=nvGreen!7,rounded corners=0.9mm,line width=0.45pt]
 (1.22,0.52) rectangle (4.98,1.34);
\node[cell,minimum width=10.5mm] at (1.9,2.11) {$P_{t-2}$};
\node[cell,minimum width=10.5mm] at (3.1,2.11) {$P_{t-1}$};
\node[cell,minimum width=10.5mm,fill=white,line width=1.0pt] at (4.3,2.11) {$P_t$};
\node[cell,minimum width=10.5mm,fill=white,line width=1.0pt] (stored0) at (6.35,2.11) {$P_t$};
\node[cell,minimum width=10.5mm] at (7.65,2.11) {$P_{t-1}$};
\draw[flow] (5.13,2.11)--(5.72,2.11);
\node[cell,minimum width=10.5mm] at (1.9,0.93) {$P_{t-1}$};
\node[cell,minimum width=10.5mm] at (3.1,0.93) {$P_t$};
\node[cell,minimum width=10.5mm,fill=white,line width=1.0pt] at (4.3,0.93) {$P_{t+1}$};
\node[cell,minimum width=10.5mm] (retained0) at (6.35,0.93) {$P_t$};
\node[cell,minimum width=10.5mm,fill=white,line width=1.0pt] at (7.65,0.93) {$P_{t+1}$};
\draw[flow] (5.13,0.93)--(5.72,0.93);
\draw[flow,draw=nvMuted!60,line width=0.5pt] (stored0.south)--(retained0.north);
\node[label,text=nvGreenDark] at (7.0,0.27) {Only the oldest slot changes};
\node[label] at (3.1,0.27) {Two cached pairs and one new pair};
\end{tikzpicture}
\endgroup}
 \caption{Bounded KV reuse. (a) Each block attends over a short span of cached key/value pairs. Because deeper blocks read hidden states that already summarize earlier frames, the receptive field in raw frames (five here, from two blocks of span three) exceeds any single span. Bounding the span keeps streaming inference identical to training. (b) Each step reads the cached pairs and the new pair $P_i=(k_i,v_i)$ (bold outline) in lag order, then overwrites the oldest slot, so the buffer is never shifted.}
 \label{fig-cache}
\end{figure}

Each block $\ell$ keeps a circular cache of fixed span and attends over the current key/value pair and the $S_\ell-1$ cached pairs, following sliding-window attention \cite{jiang2023mistral}. Each step overwrites the oldest pair, and reading in lag order rather than buffer position recovers temporal order without shifting anything. Short spans suffice because deeper blocks read hidden states that already summarize earlier frames, and the receptive field in raw frames compounds across blocks,

\begin{equation}
 R=1+\sum_{\ell=1}^{L}(S_\ell-1).
 \label{eq-cone}
\end{equation}
Two blocks of span three, for example, see five frames while each caches only two (Fig.~\ref{fig-cache}a). Each environment has its own cache and resets asynchronously without mixing histories.

For batch $B$, key/value width $d_h$, and $n_q$ query heads, cache storage is $O(Bd_h\sum_\ell(S_\ell-1))$ and attention work per step is $O(Bn_qd_h\sum_\ell S_\ell)$, and neither grows with rollout length. A fused IO-aware kernel \cite{dao2022flashattention,tillet2019triton} reads the circular cache, computes attention, and writes the new pair in one operation, and CUDA graph replay removes most remaining launch overhead, the dominant per-step cost identified in Sec.~\ref{sec-arch}. Training uses the same spans and lag biases, and streaming and full-window evaluation agree to floating-point roundoff. Per-step cost is that of one new frame rather than the whole window, with identical model predictions.

\subsection{Variable Contact Handling}
\label{sec-variable-method}

Learning contact is hard because contact forces are stiff and discontinuous \cite{parmar2021stiff}. Prior learned models have addressed this with structured contact geometry and Jacobians \cite{pmlr-v155-pfrommer21a}. NeRD avoids much of the difficulty by keeping collision detection analytical and giving the network contact geometry as input, but only at fixed probes. The variable-contact encoder keeps the analytical detection and drops the probes. At each predicted state the collision detector is run, and whatever contacts it reports form $\mathcal C_t$. Because their number and order change from step to step, they are encoded as an unordered set rather than assigned to fixed slots.

Each detected contact contributes one feature row for each robot link involved. Contacts involving only external objects are excluded, since they cannot affect the robot's next state. A row identifies the bodies, via a learned embedding for robot links and a static/dynamic flag for external ones, and gives the local geometry and motion in the robot frame, namely position, normal, lever arm, signed gap, and relative surface velocity. The $n_t$ rows are padded to a capacity $K$ chosen to cover the expected number of contacts.

To give the model a fixed-size input, each valid row $c_{t,j}$ is encoded by a shared network $\phi$ into $e_{t,j}=\phi(c_{t,j})\in\mathbb R^D$ and pooled into $M$ summary slots \cite{zaheer2017deepsets,pmlr-v97-lee19d},
\begin{equation}
 b_{t,m}=\sum_{j=1}^{n_t}
 \frac{\exp(a_m^\top e_{t,j})}
 {\sum_{r=1}^{n_t}\exp(a_m^\top e_{t,r})}\,e_{t,j},
 \quad m=1,\ldots,M,
 \label{eq-contact}
\end{equation}
where $a_m$ is a learned scoring vector. Each slot is a soft weighted average over all contacts, so the result is independent of their order and summarizes the observed set rather than a fixed location on the robot. The slots are flattened into the bank $B_t$ and paired with the normalized count $\rho_t=\log(1+n_t)/\log(1+K)$, which restores what averaging discards when several similar contacts occur. An empty set gives a zero bank and count. Pooling costs $O(KMD)$ and needs no contact-to-contact attention, adding little per-step work.

The pair $(B_t,\rho_t)$ enters through the three $G$ terms of Sec.~\ref{sec-arch}, implemented as gated bias-free linear maps. At the input embedding, contact shapes the history representation. At the readout and increment, the current contacts act directly on the predicted state, since contact forces act within the current step and should not be diluted by history. Detection is re-run at every predicted state, while cached hidden states carry the effect of earlier contacts without storing their rows. Contacts can therefore appear anywhere on the robot with no redesign of probes.

\subsection{Two-Way Coupling via Hybrid Simulation}
\label{sec-coupling-method}

A finger pushing a block must move it and be pushed back. NeRD predicts only the robot's state, so an object simulated alongside it can be pushed but cannot resist. Imposing the predicted finger trajectory moves the block with no reaction on the finger. Two-way coupling restores that reaction.

\begin{figure}[t]
 \centering
 \resizebox{\columnwidth}{!}{\begingroup
\definecolor{nvGreen}{HTML}{76B900}
\definecolor{nvGreenDark}{HTML}{4C8500}
\definecolor{nvInk}{HTML}{1A1D1A}
\definecolor{nvMuted}{HTML}{626862}
\definecolor{nvPanel}{HTML}{F4F6F2}
\definecolor{nvGreenFill}{HTML}{EDF6E2}
\begin{tikzpicture}[
 x=1cm,y=1cm,font=\sffamily\scriptsize,line cap=round,line join=round,
 box/.style={draw=nvInk!52,fill=white,rounded corners=1.15mm,line width=0.62pt,align=center,inner xsep=1.3mm,inner ysep=1.05mm},
 learned/.style={box,draw=nvGreenDark,fill=nvGreenFill,line width=0.72pt},
 analytic/.style={box,draw=nvInk!58,fill=nvPanel},
 flow/.style={-{Stealth[length=1.38mm,width=1.08mm]},draw=nvInk!78,line width=0.64pt,shorten <=0.5pt,shorten >=0.9pt,rounded corners=1mm},
 response/.style={flow,draw=nvGreenDark,line width=0.74pt},
 note/.style={fill=white,text=nvMuted,inner sep=1pt,align=center},
 badge/.style={circle,fill=nvGreenDark,text=white,minimum size=3.2mm,inner sep=0pt,font=\sffamily\tiny\bfseries}]
\path[use as bounding box] (0,0) rectangle (8.8,7.5);
\node[box,text width=5.7cm,minimum height=0.8cm] (input) at (4.4,6.95)
 {\textbf{Synchronized physical state}\\Robot, objects, and applied control};
\node[learned,text width=2.55cm,minimum height=1.1cm] (robot) at (2.55,5.38)
 {\textbf{FlashNeRD}\\Robot free motion\\$a_{r,t}^{\mathrm{free}}=\Delta\hat v_t^{\mathrm{free}}/h$};
\node[analytic,text width=2.55cm,minimum height=1.1cm] (object) at (6.25,5.38)
 {\textbf{Newton}\\Object free dynamics\\$a_{o,t}^{\mathrm{free}}$};
\node[analytic,text width=5.7cm,minimum height=2.18cm] (solve) at (4.4,3.30) {};
\node[font=\sffamily\scriptsize\bfseries] at (4.4,4.02) {Shared Newton constraint solve};
\draw[draw=nvGreenDark,line width=2pt] (2.88,2.96)--(3.30,3.35)--(3.75,3.08)--(4.38,3.08);
\foreach \x/\y in {2.88/2.96,3.30/3.35,3.75/3.08}
 {\filldraw[draw=nvGreenDark,fill=white,line width=0.65pt] (\x,\y) circle[radius=0.075];}
\path[draw=nvInk!65,fill=white,rounded corners=0.4mm,line width=0.75pt]
 (4.39,2.85) rectangle (5.19,3.52);
\fill[nvGreenDark] (4.39,3.08) circle[radius=0.055];
\draw[response,line width=0.85pt] (4.31,3.69)--(3.68,3.69);
\draw[flow,line width=0.85pt] (4.47,3.69)--(5.10,3.69);
\node[text=nvGreenDark,font=\sffamily\tiny] at (3.48,2.64) {Robot};
\node[text=nvMuted,font=\sffamily\tiny] at (4.79,2.64) {Object};
\node[text=nvMuted,font=\sffamily\tiny] at (4.4,2.36) {Contact reactions and joint constraints};
\node[learned,text width=5.7cm,minimum height=0.93cm] (integrate) at (4.4,1.02)
 {\textbf{Integrate both states}\\Resolved robot and object positions and velocities};

\coordinate (fork) at (4.4,6.22);
\draw[flow] (input.south)--(fork);
\draw[response] (fork)--(2.55,6.22)--(robot.north);
\draw[flow] (fork)--(6.25,6.22)--(object.north);
\fill[nvInk!70] (fork) circle[radius=0.6pt];
\draw[response] (robot.south)--(2.55,4.39);
\draw[flow] (object.south)--(6.25,4.39);
\draw[response] (solve.south)--node[note,right=0.8mm] {Motion after interaction} (integrate.north);
\draw[response] (integrate.west)--(0.48,1.02)--(0.48,6.95)--(input.west);
\node[note,rotate=90,text=nvGreenDark] at (0.48,3.63) {Robot feedback before the next prediction};
\draw[flow] (integrate.east)--(8.32,1.02)--(8.32,6.95)--(input.east);
\node[note,rotate=90] at (8.32,3.63) {Object state for the next physics step};
\node[badge] at (robot.north west) {1};
\node[badge] at (solve.north west) {2};
\node[badge] at (integrate.north west) {3};
\end{tikzpicture}
\endgroup}
 \caption{Two-way coupling. FlashNeRD predicts free robot motion. MJWarp supplies analytical object dynamics, solves contacts and joint constraints for robot and objects jointly, and integrates both. Object reactions act on the robot, whose resolved state feeds the next prediction.}
 \label{fig-coupling}
\end{figure}
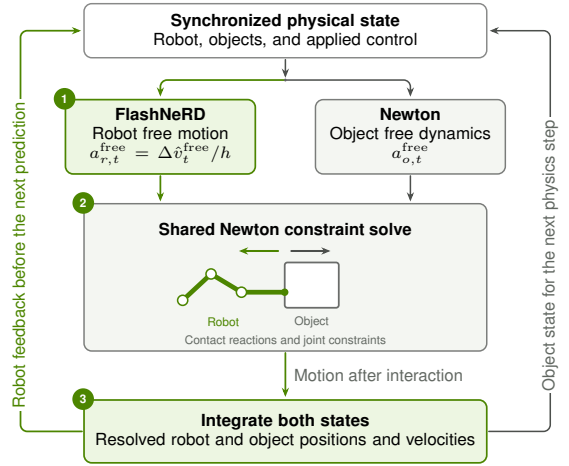

Hybrid schemes that pair learned components with differentiable simulation \cite{heiden2021neuralsim} or analytical constraints \cite{pmlr-v168-jiang22a} motivate the split adopted here. The network predicts robot motion under motor torques \emph{before} contact with objects, while MJWarp simulates the objects with its analytical dynamics and performs the contact and constraint solve for robot and objects together. The resolved robot state, not the network's free prediction, is the input to the next step (Fig.~\ref{fig-coupling}). Unlike the standalone models of Secs.~\ref{sec-arch} to \ref{sec-variable-method}, whose targets include contact effects, the coupled model learns pre-contact dynamics only, since contact effects come from the solve and learning them would count them twice.

For a fixed-base robot with positions $q_t$, velocities $v_t$, and applied torques $\tau_t$ (position targets first become torques), the model predicts a pre-constraint velocity increment,
\begin{equation}
 \Delta\hat v_t^{\mathrm{free}}=F_\theta^{\mathrm{free}}(q_t,v_t,\tau_t),
 \qquad
 a_{r,t}^{\mathrm{free}}=\Delta\hat v_t^{\mathrm{free}}/h,
 \label{eq-free-motion}
\end{equation}
where $h$ is the physics step. Training targets are velocity increments before contact, joint constraints, and integration.

Let $Q_t$ stack robot and object coordinates and $a_t^{\mathrm{free}}$ combine Eq.~\eqref{eq-free-motion} with the objects' analytical free acceleration. With the articulated mass matrix $M(Q_t)$, including armature \cite{featherstone1983calculation}, and constraint Jacobian $J(Q_t)$, MJWarp solves
\begin{equation}
 M(Q_t)a_t=M(Q_t)a_t^{\mathrm{free}}+J(Q_t)^\top f_t
 \label{eq-coupled-response}
\end{equation}
for constraint forces $f_t$ (contacts, joint limits, self-contacts) and constrained acceleration $a_t$, then integrates velocity and position. Since the learned increment precedes the solve, contact forces accelerate the object and decelerate the robot.

Geometry, inertia, and constraints thus remain analytical while only the robot's free dynamics are learned, and robot and objects advance on a common timestep. The interface assumes a fixed base. A floating base would additionally require predicting base motion.

\section{Results}
\label{sec-results}

NeRD has already shown zero-shot sim-to-real transfer and fine-tuning on real data \cite{pmlr-v305-xu25g}, so the experiments below focus on what FlashNeRD adds. Section~\ref{sec-accuracy} measures the transition model alone, Secs.~\ref{sec-rl} and \ref{sec-variable-results} place it inside policy learning and planning, and Sec.~\ref{sec-coupling-results} couples it to an analytically simulated object.

\subsection{Model-Only Performance and Prediction Accuracy}
\label{sec-accuracy}

We show that \textbf{FlashNeRD increases model-only inference throughput while improving prediction accuracy} over recursive rollouts, where predictions become inputs and errors compound. Parameter-matched Optimized NeRD and \method{} share data, splits, controls, ten-frame history, normalization, and evaluation trajectories on CartPole, Franka, and G1. CartPole and Franka data were generated with Newton's SolverFeatherstone; G1 used MJWarp. CartPole uses about 9k training trajectories and 2.7M parameters, versus 90k and 10.9M for Franka and G1. Training uses Adam, global batch 4,096, unit gradient clipping, FP16 autocast, and FP32 parameters, optimizer state, and loss, with 5M updates for CartPole and 3M for Franka and G1.

\begin{figure}[t]
 \centering
 \includegraphics[width=0.85\columnwidth]{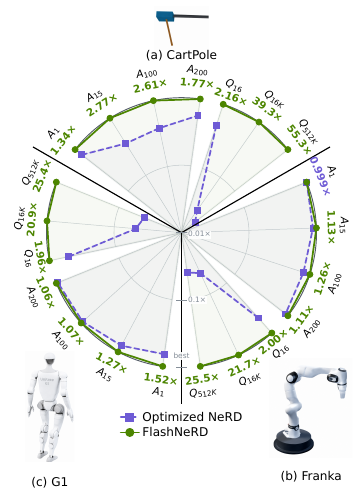}
 \caption{Model-only throughput and prediction accuracy on CartPole, Franka, and G1. Each spoke shows reciprocal rollout NRMSE $A_H=1/E_H$ at $H=1,15,100,200$ or throughput $Q_B$ at batch sizes 16, 16{,}384, and 512{,}000. Values are on a logarithmic radial scale normalized by the better model on each spoke, so a larger radius is better. Rim labels give the FlashNeRD-to-Optimized-NeRD ratio, colored by the favored method.}
 \label{fig-main-comparison}
\end{figure}

Optimized NeRD retains NeRD's architecture but pads relevant matrix dimensions to multiples favored by FP16 Tensor Core kernels \cite{nvidia2023matmul}. Both implementations use shape-specialized kernels and CUDA graph replay, with linear projections in FP16 and model parameters, residual additions, and normalization in FP32. For throughput, these shared optimizations focus the comparison on architectural differences. \method{} uses four parallel blocks with spans $[4,3,3,3]$ and one width-64 key/value head, processing one new frame per step after priming the cache with the initial history. G1 uses eight fixed probes; variable contact is evaluated in Sec.~\ref{sec-variable-results}. Figure~\ref{fig-main-comparison} normalizes comparisons per spoke, while Table~\ref{tab-prediction-accuracy} reports absolute values.

We measure accuracy by prefix-averaged rollout NRMSE, the root-mean-square per-channel prediction error normalized by training-state standard deviations and averaged over channels, steps through $H$, and trajectories. G1 root-orientation scales use canonical rotation vectors. Starting from ground-truth history, 2,048 trajectories per system are evaluated at $H\in\{1,15,100,200\}$. $H=1$ measures the first open-loop prediction, while larger $H$ averages accumulated error.

Throughput at batch $B$ is $Q_m(B)=B/[10^3L_m(B)]$ million transitions/s, where $L_m(B)$ is the median per-step latency in milliseconds over nine steady-state replay groups on one NVIDIA GB200. Model-only timings use pre-normalized, padded inputs already stored in GPU memory. Optimized NeRD processes ten frames per step, versus one for \method{}.

\begin{table}[t]
 \centering
 \caption{Recursive-rollout NRMSE and model-only throughput.}
 \label{tab-prediction-accuracy}
 \footnotesize
 \setlength{\tabcolsep}{1.6pt}
 \renewcommand{\arraystretch}{1.04}
 {\itshape (a) Prefix-averaged rollout NRMSE. Lower is better.}\par\vspace{2pt}
 \begin{tabular*}{\columnwidth}{@{\extracolsep{\fill}}llcccc@{}}
  \toprule
  \textbf{System} & \textbf{Model} & $\boldsymbol{E_1}$ & $\boldsymbol{E_{15}}$ & $\boldsymbol{E_{100}}$ & $\boldsymbol{E_{200}}$\\
  \midrule
  \multirow{2}{*}{CartPole} & Opt. NeRD & $5.89{\times}10^{-5}$ & $4.78{\times}10^{-4}$ & $8.43{\times}10^{-3}$ & $0.106$\\
   & \method{} & $\mathbf{4.41{\times}10^{-5}}$ & $\mathbf{1.73{\times}10^{-4}}$ & $\mathbf{3.23{\times}10^{-3}}$ & $\mathbf{0.0599}$\\
  \addlinespace[2pt]
  \multirow{2}{*}{Franka} & Opt. NeRD & $\mathbf{5.005{\times}10^{-4}}$ & $5.63{\times}10^{-3}$ & $0.291$ & $0.747$\\
   & \method{} & $5.008{\times}10^{-4}$ & $\mathbf{5.00{\times}10^{-3}}$ & $\mathbf{0.232}$ & $\mathbf{0.674}$\\
  \addlinespace[2pt]
  \multirow{2}{*}{G1} & Opt. NeRD & $0.0351$ & $0.0847$ & $0.247$ & $0.319$\\
   & \method{} & $\mathbf{0.0231}$ & $\mathbf{0.0667}$ & $\mathbf{0.230}$ & $\mathbf{0.302}$\\
  \bottomrule
 \end{tabular*}
 \vspace{4pt}

 {\itshape (b) Throughput ($10^6$ transitions/s). Higher is better.}\par\vspace{2pt}
 \begin{tabular*}{\columnwidth}{@{\extracolsep{\fill}}llccc@{}}
  \toprule
  \textbf{System} & \textbf{Model} & $\boldsymbol{Q_{16}}$ & $\boldsymbol{Q_{16{,}384}}$ & $\boldsymbol{Q_{512{,}000}}$\\
  \midrule
  \multirow{2}{*}{CartPole} & Opt. NeRD & $0.0805$ & $1.22$ & $1.22$\\
   & \method{} & $\mathbf{0.174}$ & $\mathbf{48.1}$ & $\mathbf{67.7}$\\
  \addlinespace[2pt]
  \multirow{2}{*}{Franka} & Opt. NeRD & $0.0697$ & $0.998$ & $0.981$\\
   & \method{} & $\mathbf{0.139}$ & $\mathbf{21.7}$ & $\mathbf{25.0}$\\
  \addlinespace[2pt]
  \multirow{2}{*}{G1} & Opt. NeRD & $0.0698$ & $0.999$ & $0.976$\\
   & \method{} & $\mathbf{0.137}$ & $\mathbf{20.9}$ & $\mathbf{24.8}$\\
  \bottomrule
 \end{tabular*}
\end{table}

\begingroup
\clubpenalty=10000
\textbf{\method{} is about $2\times$ faster at batch 16}, where the underused GPU is bound by sequential launches that the parallel block roughly halves, \textbf{and $25$ to $55\times$ faster at batch 512{,}000}, where arithmetic dominates and cached history allows it to process only the new frame rather than the full ten-frame history. Rollout \textbf{error falls by 25.2\% to 63.8\% on CartPole and 5.4\% to 34.2\% on G1}. Franka ties at $H=1$ (0.06\% difference), with 9.8\% to 20.5\% reductions at larger $H$. Accuracy is preserved and usually improved.

CartPole ablations at $H=500$ show comparable NRMSE across activations (SwiGLU, GEGLU, ReGLU, GLU, MishGLU), normalization (RMSNorm, LayerNorm), and precision (FP32, FP16, BF16), supporting the architectural explanation for accuracy.
\endgroup

\subsection{Locomotion Learning with Fixed Contact}
\label{sec-rl}
We test \textbf{whether faster transitions shorten end-to-end policy training}, including PPO updates and environment overhead. ANYmal-C velocity tracking uses PPO \cite{schulman2017ppo} in a shared tensor environment, measuring complete training time as in massively parallel locomotion learning \cite{rudin2022minutes}.

All learned models use fixed contact with the same four foot-sphere probes, which matches NeRD's formulation and isolates the architectural and inference changes from variable contact. Four dynamics backends are compared. NeRD uses its full-window transformer. Optimized NeRD adds CUDA graph replay and TF32 matrix multiplication with FP32 weights and state. \method{} uses the streaming architecture. MJWarp within Newton is the analytical reference.

We collect task-independent MJWarp data by randomizing base poses, base velocities, and joint configurations. Joint-position target offsets are sampled uniformly within $\pm0.75$\,rad of the default pose every 20\,ms. Rollouts last up to 100 steps, retaining falls but dropping invalid and reset transitions. States, targets, and probe features form supervised transitions capturing the ANYdrive actuator response. With no walking policy or reward involved, the frozen model can serve other RL objectives with the same actuators.

Both architectures share this dataset, a 5\% trajectory validation split, normalization, 60,000 training updates, and a 13-frame receptive field, with 8.88M parameters for \method{} and 9.81M for NeRD. PPO runs on one GB200 with 16,384 environments, 300 iterations, 24 transitions per environment per iteration, identical settings, and one training seed per method. Timings exclude data collection, dynamics training, and process initialization.

\begin{figure}[t]
 \centering
 \includegraphics[width=\columnwidth]{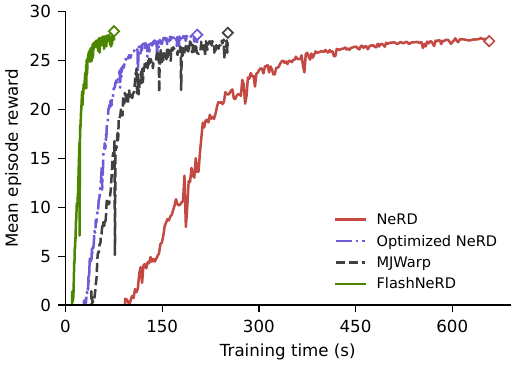}
 \caption{PPO training reward versus accumulated training time (sum of logged iteration durations) for ANYmal-C velocity tracking with 16,384 environments over 300 iterations, with fixed contact for all learned models. Diamonds mark final policies evaluated in MJWarp with mean actions and the training reward, showing faster training reaches comparable quality.}
 \label{fig-rl-reward}
\end{figure}

A training-reward threshold is set at 95\% of MJWarp's mean reward over its final 20 iterations. The first unsmoothed crossing occurs at 34.3\,s with \method{}, 93.2\,s with Optimized NeRD, 129.9\,s with MJWarp, and 357.5\,s with NeRD (Fig.~\ref{fig-rl-reward}). Including training-loop overhead, totals are 79.1, 204.9, 255.9, and 657.5\,s in the same order, \textbf{making \method{} 2.6$\times$ faster than Optimized NeRD and 3.2$\times$ faster than the analytical simulator}.

Relative to Optimized NeRD, collection latency drops from 22.51 to 4.41\,ms per step and network latency from 19.63 to 1.32\,ms, after which contact processing and policy inference become the largest collection costs. PPO updates stay near 0.11 to 0.13\,s per iteration. The dynamics model is no longer the bottleneck, and further speedup is capped by the rest of the loop.

\begin{figure}[t]
 \centering
 \includegraphics[width=\columnwidth]{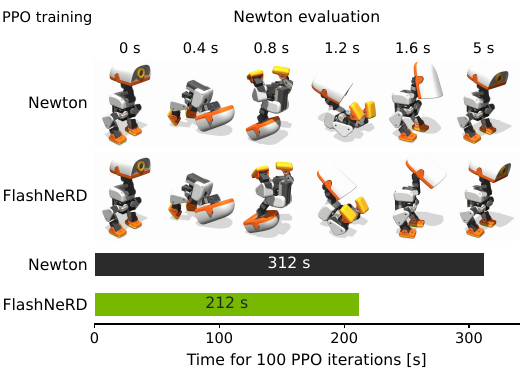}
 \caption{Microduck forward rolling with PPO fine-tuning in Newton and FlashNeRD, both evaluated in Newton. Bars show mean wall time for 100 PPO iterations over two seeds.}
 \label{fig-microduck}
\end{figure}

Final policies are evaluated in MJWarp over 256 environments and 1,000 steps with mean actions and the training reward. Rewards are 28.0 for \method{}, 27.6 for Optimized NeRD, 27.0 for NeRD, and 27.8 for MJWarp training. Faster training therefore preserves policy quality in MJWarp. Microduck forward rolling provides another illustrative locomotion example (Fig.~\ref{fig-microduck}).

\begin{figure*}[t]
 \centering
 \includegraphics[width=0.9\textwidth]{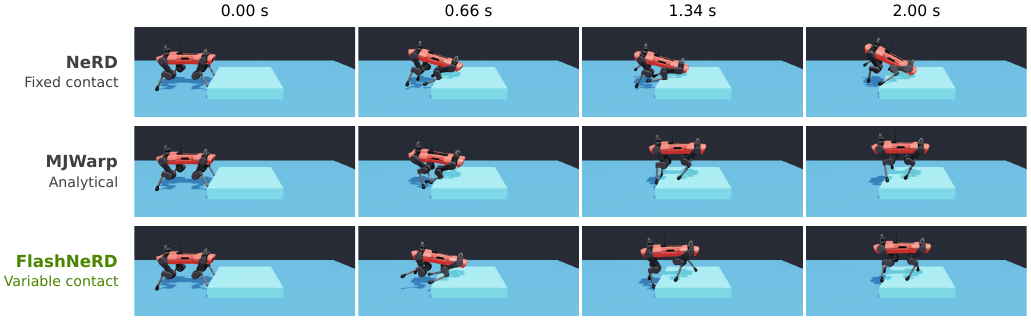}
 \caption{DIAL-MPC climbing a 0.181\,m platform. Rows are the planning dynamics and columns are matched times. All selected actions are executed in MJWarp within Newton, so only the planner's model differs. The robot climbs with MJWarp and with variable-contact FlashNeRD, and fails to clear the edge with fixed-contact NeRD.}
 \label{fig-dial-rollouts}
\end{figure*}

\subsection{Variable Contact for Contact-Rich Planning}
\label{sec-variable-results}

We next test \textbf{whether variable contact improves control when contacts occur outside predefined probe locations}. During climbing, the platform edge contacts different parts of the feet and shins as the robot moves, making sparse fixed probes insufficient to represent all relevant contacts. We use DIAL-MPC \cite{xue2025dial}, a sampling-based MPC controller, because its thousands of dynamics queries per decision test both contact fidelity and rollout speed.

DIAL-MPC samples 160 action trajectories over a 24-step horizon for ANYmal-C, refines them with a shared objective, and executes the chosen action in MJWarp. Only the planning dynamics differ across methods. Each trial lasts 100 control frames at 20\,ms. Six terrains vary platform distance and height, with heights from 0.150 to 0.193\,m.

Fixed-contact NeRD is given 20 probes covering feet and shins against ground and platform, so it knows the platform geometry but only at those points. \method{} consumes Newton's detected contacts with capacity 64 pooled into 16 slots, with detection run at every predicted state.

A task-independent dataset is collected in MJWarp by randomizing platform height and distance, robot states on the ground, edge, and top, and joint-position targets. Targets within $\pm0.6$\,rad of the nominal pose follow an autoregressive process with coefficient 0.9 for up to 100 steps. Falls and both contact representations are recorded at each starting state, and no planner, expert, or reward guides actions. Both models train for 100,000 updates with the same optimizer schedule, the same normalization scheme with separately fitted statistics, and a 5\% trajectory holdout, remaining frozen during planning. Because architecture and contact representation change together, the experiment evaluates the system rather than contact alone.

A climb succeeds when the base stays at least 0.12\,m past the edge and 0.45\,m above the platform, with body-up vertical component at least 0.85, for eight consecutive frames, which requires sustained upright placement beyond the edge. Table~\ref{tab-climb} reports success and latency.

\begin{table}[t]
 \centering
\caption{DIAL-MPC on six terrains with execution in Newton (MJWarp). Planning time is the median per control decision.}
\label{tab-climb}
 \small
 \setlength{\tabcolsep}{4pt}
 \begin{tabular}{lcc}
  \toprule
  Rollout dynamics & Climbs & Planning time (ms)\\
  \midrule
  NeRD with fixed contact & 1/6 & 2441\\
  MJWarp & 6/6 & 3528\\
  \method{} with variable contact & 6/6 & 1720\\
  \bottomrule
 \end{tabular}
\end{table}

The robot \textbf{completes all six climbs with \method{} and with MJWarp, and only one with fixed-contact NeRD}. Figure~\ref{fig-dial-rollouts} shows a 0.181\,m example. \textbf{Median planning time falls by 1.42$\times$ relative to NeRD and 2.05$\times$ relative to MJWarp}. One-step root-position increment RMSEs logged during training are 0.541\,mm for \method{} and 0.813\,mm for NeRD. They use different validation subsets and only indicate rather than strictly compare accuracy.

During the illustrated MJWarp climb, collision detection produces 11 to 40 contact rows per step while 3 to 16 fixed probes are active. All rows fit within capacity. Both models see the same physics, but the variable set follows the geometry over the edge, enabling all six climbs at half the analytical simulator's planning time. Planning still takes seconds per decision, above the 20\,ms control interval, making this an offline or slowed-clock result.

\subsection{Coupled Manipulation}
\label{sec-coupling-results}
We test \textbf{whether a learned hand can manipulate an object absent from training} through Allegro-hand cube reorientation, with the cube and all contacts simulated analytically in MJWarp. One-way and two-way coupling use the same frozen model and differ only in whether reaction forces from the cube are applied to the hand.

\begin{figure}[t]
 \centering
 \includegraphics[width=\columnwidth]{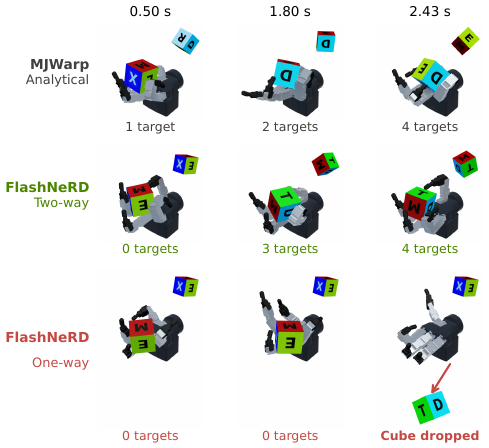}
 \caption{Cube reorientation in MJWarp for policies trained with different dynamics (rows). Small cubes show target orientations and counts show completions. The policy trained with one-way coupling never felt the cube resist during training and drops it (arrow), while the two-way FlashNeRD and MJWarp-trained policies complete four targets in the same time.}
 \label{fig-allegro-rollouts}
\end{figure}

Allegro free dynamics are learned from 4.19 million random state-command samples with 5\% held out. Positions mix uniform samples and reset-pose perturbations, velocities mix Gaussian and uniform samples, and random position targets are converted to motor torques. MJWarp computes pre-constraint velocity increments for the Allegro hand without the cube at 240\,Hz, providing supervision for FlashNeRD.

In two-way training, \method{} predicts the hand's free velocity increment, and MJWarp advances the cube and jointly solves contacts and joint constraints. The resolved hand state feeds the next prediction. One-way coupling retains hand joint limits and self-contacts but discards cube reactions. PPO observes the cube pose in both cases.

Each method uses one policy seed, random actor and critic initialization, identical PPO settings, and 5,501 updates of 16 policy steps per environment (262,144 transitions per update). We select the best of nine shared checkpoints by mean completed targets over 256 MJWarp validation episodes each. Final evaluation uses a separate seed and 512 episodes per policy, including failures. \emph{All final evaluations and displayed frames simulate both hand and cube analytically in MJWarp}. Labels identify training dynamics.

Episodes last up to 10\,s. We count a target when orientation error falls below 0.2\,rad, then sample another orientation. Policies act at 30\,Hz and physics at 240\,Hz. In Fig.~\ref{fig-allegro-rollouts}, the one-way policy drops the cube at 2.47\,s without completing a target, while the other two complete four by 2.43\,s.

\textbf{Two-way \method{}, MJWarp, and one-way coupling average 14.26, 14.49, and 1.79 targets per episode}, with 95\% bootstrap intervals of $[13.96, 14.54]$, $[14.19, 14.78]$, and $[1.65, 1.93]$ from resampling environments while keeping each environment's two episodes together. At least two targets are completed in 99.0\%, 98.2\%, and 49.2\% of episodes, respectively, with cube-drop rates of 9.0\%, 8.2\%, and 18.9\%.

Discarding cube reactions, all else fixed, reduces completed targets eightfold and doubles the drop rate, showing that nearly all the two-way policy's manipulation ability depends on cube reactions during training. With reactions, \textbf{policies perform within 2\% of MJWarp-trained PPO} in MJWarp.

\section{Limitations}
\label{sec-limitations}
Each model is specific to a robot, actuator model, and timestep. Bounded history may miss unobserved dynamics with longer memory, and variable contacts must fit a preset capacity. Analytical collision detection limits end-to-end speedup, while coupled simulation additionally requires analytical constraint resolution and a fixed base. Our timings use GB200 GPUs, and performance on smaller GPUs or onboard hardware remains untested.

\section{Conclusion}
FlashNeRD combines streaming inference, variable contact, and two-way coupling for fast, contact-rich robot dynamics. Its model is up to $55\times$ faster than Optimized NeRD with lower rollout error, enables climbing where fixed probes fail, and trains cube-reorientation policies within 2\% of simulator-trained policies in completed targets per episode. These results support learned simulation for policy learning and planning, with real-world adaptation offering a path toward dynamics models that improve with experience.

\bibliographystyle{ieeetr}
\bibliography{references}
\end{document}